# From locomotion to cognition

# Table of contents

# 1 Summary of results in last project period (1. 10. 2009 – 30. 9. 2010)

## (a) Work package 1: On-line manipulation of morphology

Our work on online manipulation of mechanical properties (wrapped in the generalized concept of morphology), described in the previous report, has opened a myriad of new questions as well as application, publication, and collaboration opportunities. Therefore, we have decided to continue investing efforts in this work package and report the advances generated with the continued activity.

**Magnetic spring**

The idea behind this actuator was to regulate the repulsion between two permanent magnets by means of a magnetic field generated by an energized coil. The effect of the magnetic field of the coil can be used in two ways. The first possibility, and the one we investigated, is to use the coil to directly reduce or increase the interaction between the permanent magnets. The second possibility is to use the coil to generate a torque on one of the permanent magnets and rotate. In this way, the two permanents magnets, originally in a repelling configuration, can be brought to an attracting, or at least less repelling configuration. Although it has been shown that the latter use of the coil can be very efficient, it requires movable parts that complicate the design of the actuator. Additionally, this second implementation of the actuator strongly relies on the geometrical relation between coil and magnets. Thus, it requires significant flexibility in the coil manufacturing process – a requirement that was not attainable in the workshop of our laboratory.

In our previous report, we identified one of the biggest problem of these actuators, namely the production of heat and the consequent demagnetization of the permanent magnet. During the experiments reported there, we opted to actively cool down the device using a ventilator mounted on top of the actuator. This solution impedes the use of the actuator on robots. To overcome this difficulty, we explored geometrical arrangements that would put the coil away from the magnets. In this way we increased the maximum allowed current on the coil, source of the thermal power, and the consequent change of the force curve of the actuator. Exploring different designs numerically, Carbajal found that the coil can be completely eliminated by the use of a ferromagnetic material affecting the field configuration (a possibility already mentioned in our previous report). Figure 1 shows the force curve of the new actuator together with the results presented in our last report, for an actuator using coils. It can be observed that the performance of the new actuator is almost the same, but with zero cost in terms of current consumption this time. The disadvantage of the new design comes form the fact that it requires a moving part: either the magnet connected to the stator or a piece of ferromagnetic material that causes the reconfiguration of the surrounding field. These results are unpublished and therefore more details are available at demand, after agreement on confidentiality.

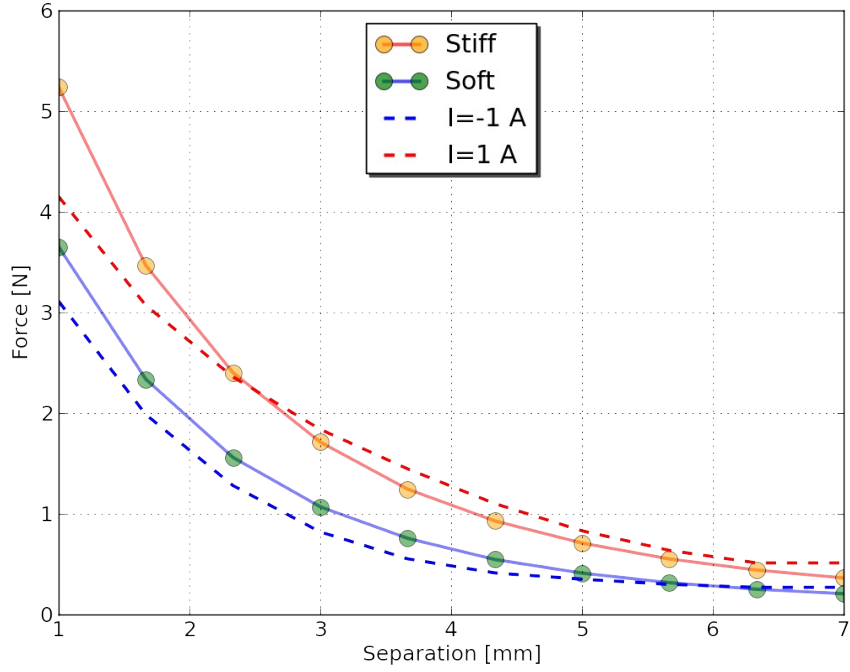

**Figure 1: Performance of new design of magnetic actuator**. Force versus displacement curves are compared. The lines with markers correspond to the new actuator and the dashed lines to the one reported previously. The upper line corresponds to the "stiff" configuration, which implies a positively magnetized coil in the old actuator. The new actuator version achieves comparable performance without spending energy.

We used the FEMM[1] software for calculating all the results presented here. The plots were done using the Python library Matplotlib[2].

*Application*

In Carbajal and Kuppuswamy (2010), the magnetic actuator was used in a feasibility study of an adaptive controller for small swimmers. There it is shown how that, though the actuation system is very simple, the intrinsic non-linearity of the actuator makes it very hard to control for the emblematic PID, designed for linear systems. Though at first sight this may look as a disadvantage of the actuation device, in a different work[3], we showed that there are controllers that exploit the non-linearities to increase their performance, for more details see WP2, section (b).

**Tunable rotary joint**

On a swimming platform, WandaX, tunable joints were used to adjust the stiffness distribution along the robot's body. The joints have been presented in our previous report. Here we extend their mathematical description. The torque generated by the joint when deflected a given angle is

$$\tau(\theta) = \frac{\left(\sqrt{r^2 + d^2 - 2\cos(\theta)\, d\, r} + r - d\right) K + F}{\sqrt{r^2 + d^2 - 2\cos(\theta)\, d\, r}} \, d\, r \sin(\theta).$$

Where r and d are geometrical factors defined by construction, K is the stiffness of the spring and F the pretension applied to it. If we approximate the function by a 3rd order polynomial expansion, we obtain

---

[1] D. C. Meeker, Finite Element Method Magnetics, Version 4.2 (15Jul2009 Mathematica Build), http://www.femm.info
[2] J. D. Hunter (2007), Matplotlib: A 2D Graphics Environment. In Computing in Science & Engineering **9(3)**, pp. 90-95
[3] AMARSi project (FP7-ICT-248311), deliverable 4.1 report.

Note that the linear component, and therefore the linear natural frequency, is independent of the stiffness of the linear spring, and varies proportional to the pretension applied. In Figure 2, we show the effect of a sudden pretension on the oscillations of the joint. After the pretension is applied, the frequency of the oscillations is considerably increased.

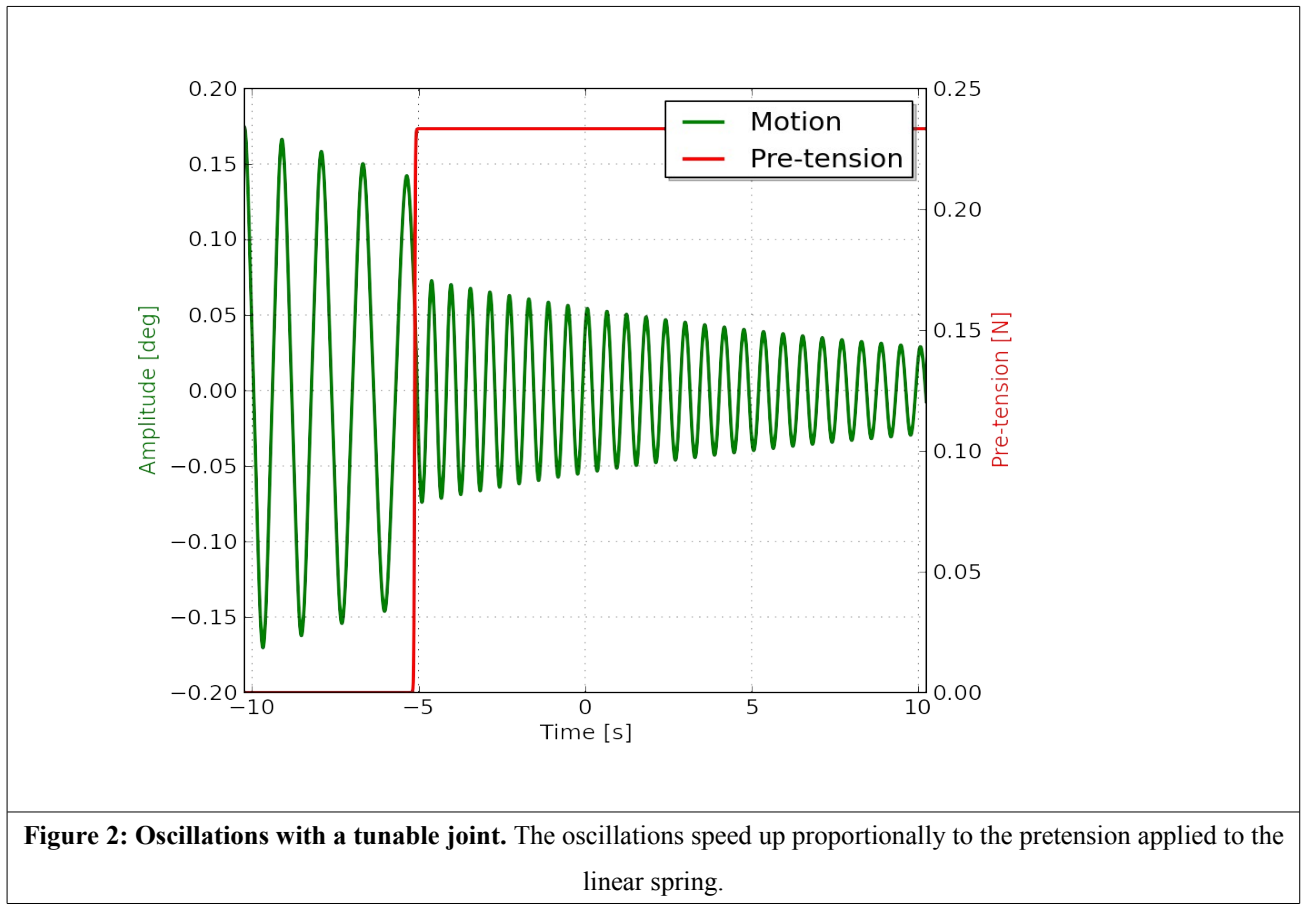

**Figure 2: Oscillations with a tunable joint.** The oscillations speed up proportionally to the pretension applied to the linear spring.

*Application*

In Ziegler et al. (2011), the authors successfully applied the tunable joint to produce a wide variety of motion patterns, providing some examples on the usability of these devices (see WP2).

**Tools of analysis**

From the preceding sections it should be noted that there are no ready-to-use tools for the analysis of systems that deal with non-linearities. In particular we would like to be able to understand how the system reacts to actuation with a characteristic frequency content (frequency spectrum). In linear systems this is known as the transfer function and one of its most remarkable use is the identification of resonances and resonant frequencies. Such equivalent does not exist in the non-linear realm, however the concepts of non-linear normal modes[4] and nonlinear output frequency response functions (NOFRF)[5] are being used with relative success. In collaboration with Dr. Z.Q. Lang from University of Sheffield, we have been studying the responses of the tunable rotatory joint for different activation signals (pretension signals), using the NOFRF method. Preliminary results are shown in Figure 3.

---

[4]A. F. Vakakis, L. I. Manevich, Yu. V. Mikhlin, V. N. Pilipchuk, A. A. Zevin (1996), Normal Modes and Localization in Nonlinear Systems. NY: Wiley, ISBN 0-471-13319-1.
[5]Z.K. Peng, Z.Q. Lang, S.A. Billings (2007), Resonances and resonant frequencies for a class of nonlinear systems, J. Sound Vibrat. **300(3-5)**, 993-1014, DOI: 10.1016/j.jsv.2006.09.012.

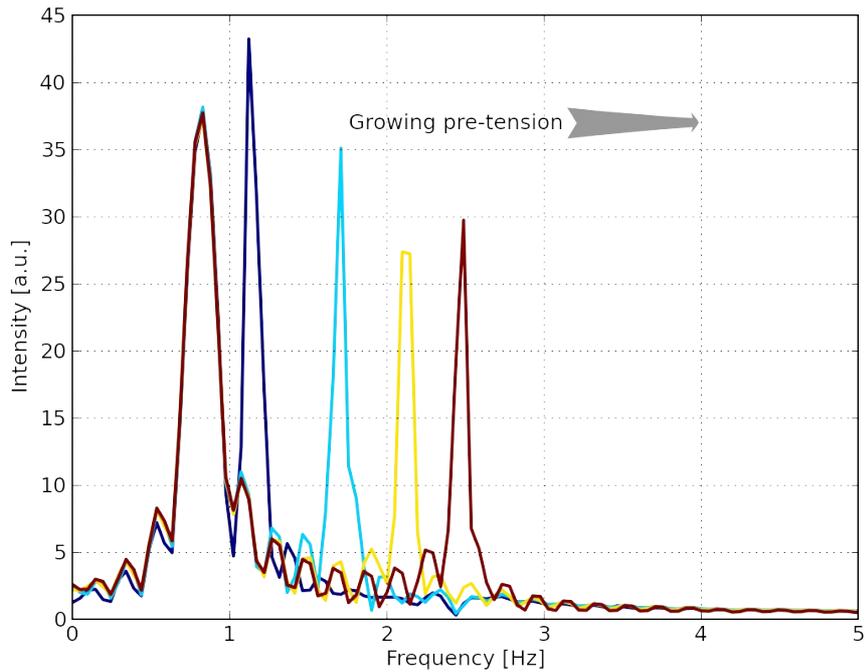

**Figure 3: Frequency spectrum of the motion of the tunable rotatory joint.** The different values of pre-tension are applied to a moving joint. The high frequencies induced on the motion are observable.

**Unsuccessful attempts and lessons learned**

As mentioned in our previous report, we explore the possibility of using the Fin Ray Effect® to alter the shape of a fin of swimming robots. Though theoretically plausible, the forces required to produce appreciable changes on 10 cm high fins, for the materials tested, are too high. The mechanism amplifies a deformation on its base (where the fin is connected to the body of the robot) at expenses of forces. Therefore, although the deformation is higher toward the end of the free structure, the exerted forces decrease. All the deformable materials tested were able to resist the forces applied with little deformation, rendering the approach unfeasible.

Additionally we attempted to induce a change of stiffness on a thin layer of Polycaprolactone, by means of controlled heating. Since the layer was meant to be used as a fin, it is required to bend, therefore the heaters had to be flexible. Flexible heaters are expensive and fragile, additionally the stiffness of the fin is changed on a rather long time scale (~10 s) and the surrounding flowing water accelerates the thermal losses. All these characteristics made the approach unattractive and the attempt was abandoned, however the possibility should not be discarded for further research.

## (b) Work package 2: Learning to exploit body dynamics

Efficient locomotion is orchestrated by the dynamic interplay of a controller, which provides actuation, and body interacting with the environment. All the components are equally important for the resulting behavior. Whereas typically it is only the controller that is subject to learning (or plasticity), we have concentrated on the role of morphology. First, we have investigated the behavior of an Adaptive Frequency Oscillator (AFO) in the role of a controller, coupled to an emulated physical system. An AFO is capable of adapting itself to the resonant frequencies of a dynamical system. If the AFO is driving the physical system at the same time, this in turn results in efficient energy transfer. In order to understand the properties of such a closed loop interaction, we have simulated the putative physical system with a dynamical system that we fully understand and whose properties we could manipulate. Second, we have engaged in experiments in swimming platforms. Taking advantage of the novel hardware solutions for fin or whole body stiffness manipulation, we have investigated how a robot can utilize this capability to improve the efficiency of its swimming (maximize thrust) and how it can expand its behavioral repertoire.

**Exploitation of nonlinear mechanical properties to boost energy transfer**

We explored numerically the performance of a single adaptive Hopf oscillator[6], a particular adaptive frequency oscillator, when strongly coupled to an oscillator with cubic nonlinearity. The former oscillator was acting as a controller driving a forcing (actuation) signal; the latter oscillator emulated the physical/mechanical system.

$$\ddot{x}+d\dot{x}+(2\pi f_0)^2 x + a_3 x^3 = \cos(\omega t),$$

where $f_0 = 3$, $a_3 = 1.20 \times 10^4$ and $\omega = \omega(t)$ is the control signal produced by the AFO with coupling parameter . It must be noted that if the coefficient of the cubic term is zero, the system is linear with a resonant frequency of 3 Hz.

The amount of mechanical energy introduced by the AFO was evaluated and compared for the linear and nonlinear system. There exists a basin of attraction where the AFO is able to approximate the resonant frequency of an input signal $\epsilon = 3 \times 10^2$ sourced from the oscillations of the linear mechanical system. However, it is unable to drive the linear system into resonance in time periods of practical interest, failing to pump energy into the system. When the AFO is connected to the system with cubic stiffness, it presents good performance and the amplitude of oscillation increases in the time window studied (~20s, considered representative for practical applications). The evolution of the amplitude of the oscillations is shown in Figure 4.

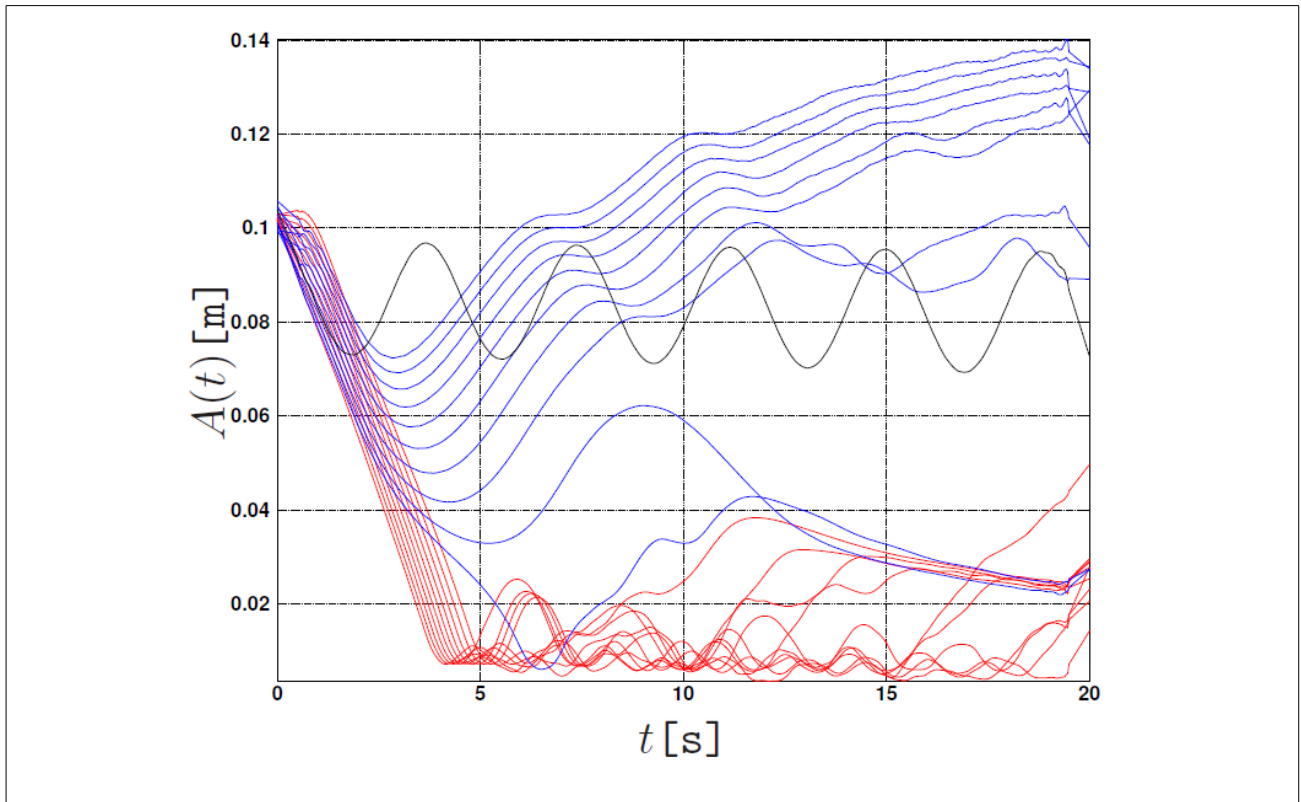

**Figure 4: Performance of AFO coupled to linear and nonlinear oscillator.** The time evolution of the amplitude of

---

[6]Ludovic Righetti, Jonas Buchli, and Auke Jan Ijspeert (2009), Adaptive frequency oscillators and applications. The Open Cybernetics and Systemics Journal **3**, pp 64–69.

> oscillations (mechanical energy) is plotted for different initial conditions. For a linear system (red) the AFO is unable to pump energy into the system and oscillations fade out. For a system with cubic nonlinearity (blue) the amplitude of the oscillations grow. In black line we show the amplitude of the linear system periodically forced at 3 Hz.

This results suggest a switch of perspective; instead of asking what controller could pump energy into the mechanical system, we can ask what mechanical system (morphology in a general sense) could exploit the properties of a controller at hand to do it. The reason for the efficient energy transfer in a non-linear case remains unknown, but we speculate that it may be related to an averaging effect caused by fast changes on the frequency of the AFO signal.

**Manipulating body stiffness in swimming robots: increasing energy efficiency and behavioral diversity**

In the swimming robots' realm, we have completed the development of the WandaX and Wanda2.0 platforms (see previous project report) and employed them in experiments on exploration and exploitation of body dynamics. They provided a key feature: the possibility to manipulate the body dynamics online. In WandaX our focus was to optimize stiffness distribution along the body to increase swimming efficiency. In Wanda2.0, the goal was to investigate how the behavioral repertoire can be expanded through tail fin stiffness manipulation.

WandaX is a swimming platform that is attached from the top and its motion is thus restricted to 2 dimensions. It consists of five segments, with a single front segment and four rear segments. A single motor is attached after the front segment; the area of the front segment is equal to the total area of all the rear segments. The only source of asymmetry in the platform – which can be responsible for swimming forward or backward - are passive compliant joints between the rear segments. We have enhanced the platform as planned with a mechanism that can manipulate the stiffness of these joints online. In Ziegler et al. (2011), in a set of experiments using online optimization, we investigated how the platform can discover optimal stiffness distribution along its body in response to different frequency and amplitude of actuation. We show that a heterogeneous stiffness distribution - each joint having a different value - outperforms a homogeneous one in producing thrust. Furthermore, different gaits emerged in different settings of the actuated joint. This work illustrates the potential of online adaption of passive body properties, leading to optimized swimming.[7]

Wanda2.0 is a freely swimming platform. Its tail fin stiffness can also be adjusted but through a different mechanism. The flexible tail fin consists of two main foils of 13 cm length and 20 cm height. In between the main foils, additional foils of different thickness and material can be inserted through an ellipse linkage. Actuation and positioning of the compliance mechanism is provided by two servo motors, one for each insertion foil. The additional foils can be inserted individually and at any length, from no insertion, leading to a soft tail fin, to fully inserted, which results in a stiffer tail fin. The enveloping main foils are not affected in terms of shape when additional foils are inserted. Therefore, the effect on swimming performance remains restricted to the stiffness only. The mechanism is shown in Fig. 5.

---

[7] More details can be obtained from the publication which we attach.

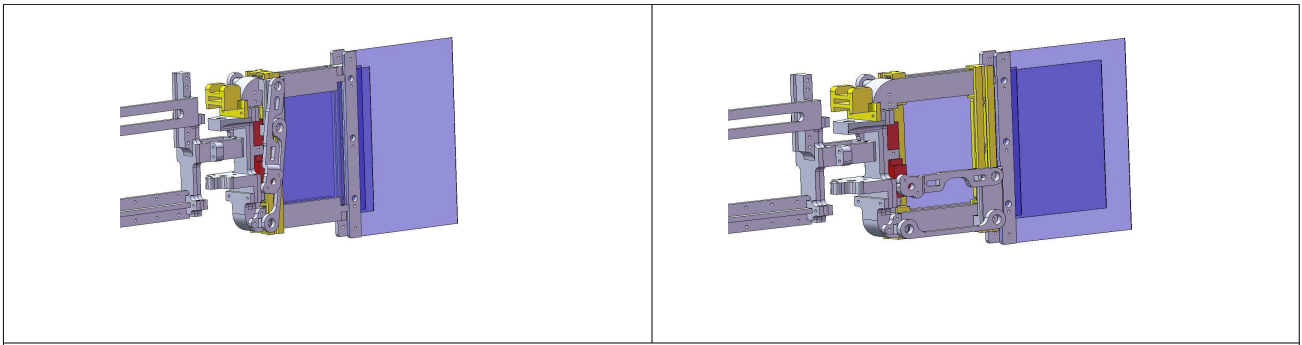

**Figure 5: Adjustable tail fin stiffness mechanism.** The compliance mechanism to soften (the foils are out, left picture) or stiffen (the foils are in, right picture) the tail fin.

As stated above, Wanda2.0 is a freely swimming platform. Therefore, it can also swim up and down in the water slope. However, it has only one motor. We have investigated, how the behavioral repertoire can be expanded by simultaneous adjustment of control parameters (amplitude, offset, and frequency for the single actuator) and body parameters (changing the tail fin stiffness). We have identified different swimming modes (or behaviors), some of them also moving actively down in the water in a corkscrew-like path (Ziegler and Pfeifer, in preparation).

Wanda2.0 platform was also equipped with a rich sensor suite (three axis accelerometer, two one axis gyroscopes, power monitor, bending sensors in tail fin, compass module, water pressure sensor). Such multimodal can be also exploited by the robot, in order to learn about its body and its action possibilities. For instance, correlations between the motor parameters, tail fin stiffness, and the water pressure sensor can be exploited to activate "gaits" that swim downward or upward. This will be the focus of our future investigation and will also relate to the work that we report in WP3.

## (c) Work package 3: Towards anticipatory behavior grounded in body dynamics

In this work package, we built on the results from the previous work packages and extended them to the cognitive realm. Given a particular action (or behavior) repertoire that a robot has learned through self-exploration (WP2), how can the behaviors be appropriately employed in different situations? The key to the selection of a suitable behavior is to be able to estimate its consequences. To this aim, we have focused on a navigation problem in our quadruped robot. The behavioral repertoire comprised a fixed number of gaits and we were interested in their consequences in terms of the change in position and orientation of the robot. We split the problem into two parts. First, we focused on the problem of path integration using self-motion cues, i.e how can a quadruped robot gauge the distance it traveled through sensor fusion, but without consulting an external reference system. Second, we addressed the problem of prediction and planning. Given that the robot has learned where and how fast can different gaits bring it (a forward model), it can use it for planning or action selection. We have used a predator-prey scenario, in which the robot needs to plan a gait sequence in order to intersect another moving robot (the prey).

**Path integration using multiple sensory modalities**

The details of the first set of experiments are described in Reinstein and Hoffmann (2011); we will only briefly summarize them here and complement with additional interpretation. Our quadruped was left to run for medium time intervals and used all the available sensory information to estimate its full body state (position, velocity, and attitude).

The sensors used were angular sensors on hips and knees, pressure sensors on feet, and an inertial measurement unit (IMU) providing linear acceleration and angular velocity signals. No external reference, such as GPS or visual landmarks, was used. Our contribution lies mainly in the combination of the sensors on the robot's legs that gives an estimate of the velocity and distance traveled, providing a legged odometer. This was then fused using an Extended Kalman Filter with the inertial measurements, giving a body state estimation. We tested our method on two different gaits, including transitions, and two terrains, one of which was highly slippery. Our solution is of interest as an application since it provides a novel solution to dead reckoning in a legged platform (so far, the focus has been mainly on wheeled robots).

At the same time, our work is interesting from a biological and cognitive science viewpoint – a perspective that was not stressed in the paper. First, the scenario we used could be further elaborated and turned into a testbed to test hypotheses regarding animal navigation, mammals or ants for instance. Second, the navigation system relies on learning the contingencies between the different sensory modalities and their relationship to velocity (or stride length). Such cross-modal associations that arise between the modalities as the robot is locomoting can be also interpreted as a form of the robot's body schema (Fig. 7 in Reinstein and Hoffmann, 2010, is an example of correlations between different sensory modalities). The pattern changes when the gait or terrain change, which can be utilized by the robot to, loosely speaking, get a "feel" for "how it is like" to run with different gaits on different substrates. We plan to pursue this direction further in the future.

**Moving target seeking with prediction and planning**

This work has been published in Oses et al. (2010) and only the key points are reported here. The scenario that we encountered in the previous section was, using probabilistic terminology, an example of filtering – computing the belief state (the posterior distribution over the current state) given all evidence to date. The state was the position and orientation of the robot. This is a useful feat for any agent (animal or robot). Another useful capability is to be able to predict future states. Such a capability is at the same time a first step toward cognition. The predictive model is a representation of the sensory-motor states which can be iterated and thus projected into the future, forming a basis for "thinking" or mental imagery. In our case, we wanted the robot to be able to predict the consequences of its actions, where, similarly to the previous section, the "consequence" was change in position and actions were gaits. After a training period of exploration or "motor babbling", the robot has learned a probabilistic forward model that allowed it to estimate where will a particular gait bring it if applied for a definite time period. For action selection, an inverse mapping to the forward model is often necessary – an inverse model. That is, which action (gait) is the best choice to achieve a particular goal (reaching a desired location). In order to pose the right challenges to our robot, we have come up with a predator-prey scenario. How does the hunter robot "catch" another mobile robot (prey)? Following the bottom-up approach to intelligence, we wanted to introduce models only when necessary (i.e. only when they significantly improve performance). Therefore, we have compared several possible architectures in the hunter robot: (a) a reactive architecture, where the hunter is using an inverse model to pick a gait that would bring him closest to the prey; (b) a prey prediction model, where a model of the prey's behavior is learned and utilized in addition; and (c) a planning model which employs the forward model to generate a gait sequence that would be needed to intersect the predicted future position of the prey. The performance measures justify the use of the predictive architectures. All the models are learned *ab initio*, without assumptions, work in egocentric coordinates, and are probabilistic in nature. We have made several simplifications in our implementation. First, we have abstracted from the problem of position sensing and prey sensing and used a GPS to emulate the prey sensing on the part of the hunter robot. Second, we used a simulated Khepera robot instead of the quadruped. We are working to alleviate these abstractions in our current work.

Nevertheless, we believe that we have preserved the most vital constraints: an embodied agent with a discrete gait repertoire and a dynamic scenario, where real-time planning is necessary.

### (d) Work package 4: Principles and dissemination

The individual scientific results, as described above, have been disseminated in the form of scientific publications. In addition, we have been working toward integrating the insights learned. To this end we have compiled a review article on body schema in robotics (Hoffmann et al., 2010), and we are working on a book chapter that deals with the implications of embodiment for behavior and cognition (Hoffmann and Pfeifer, in preparation).

The article on body schema surveyed the body representations in biology from a functional or computational perspective to set ground for a review of the concept of body schema in robotics. First, we examined application-oriented research: how a robot can improve its capabilities by being able to automatically synthesize, extend, or adapt a model of its body. Second, we summarized the research area in which robots are used as tools to verify hypotheses on the mechanisms underlying biological body representations. We identified trends in these research areas and proposed future research directions[8].

In addition to dissemination in the form of scientific publications, we have also targeted the general public through presence on various fairs, and through diverse media presence (press, media) – see the section "Dissemination and special events". At the same time, we have been active in establishing a connection to the area of education. Please refer to the section "Impact on education" under "Overview of the results of the entire project".

## 2     Overview of contributions of SNF researchers

### Funded by the project

Juan Pablo Carbajal – Juan Pablo Carbajal was mainly active in WP1 (On-line manipulation of morphology),WP 2 (exploitation of body dynamics), and collaborated in WP 3 (Towards anticipatory behavior grounded in body dynamics). He was involved in the swimming platforms and also in test beds for the actuators.

Matej Hoffmann – Matej Hoffmann was mainly active in WP 2 (exploitation of body dynamics), WP 3 (Towards anticipatory behavior grounded in body dynamics), and WP 4 (Principles and dissemination). He was involved in the quadrupedal and also the swimming platforms.

### Supervision of MSc., BSc., theses

Elias Hagmann (MSc. thesis, A simplified approach towards legged locomotion control, supervision: M. Hoffmann, J. P. Carbajal, M. Lungarella)

## 3     Cooperation

*Department of Measurement, Faculty of Electrical Engineering, Czech Technical University in Prague*

This cooperation was realized through a 6-month internship of Michal Reinstein at the AI Lab in Fall 2009. Through this cooperation, we have been able to bring in the expertise in Kalman filtering, and autonomous navigation systems, which was applied to the quadrupedal platform. Please refer to section "Path integration using multiple sensory modalities" and to Reinstein and Hoffmann (2011).

---

[8]Please see the attached publication for details.

*Fatronik-Tecnalia*

Fatronik-Tecnalia is a technology centre based in San Sebastian, Spain. We have established a cooperation with the Neuroengineering department through a 5-month internship of Dr. Noelia Oses at our lab during Spring 2009. Noelia Oses was working jointly with M. Hoffmann and J.P. Carbajal on anticipatory behavior in the quadrupedal platform. Please consult section "Moving target seeking with prediction and planning" and Oses et al. (2010).
http://www.fatronik.com/en/index.php

*Department of Automatic Control and Systems Engineering, University of Sheffield*

Dr. Zi-Qiang Lang is a senior lecturer in systems and control engineering interested in the analysis and design of nonlinear systems in the frequency domain. There is an ongoing collaboration with one of his PhD students, Rafael Bayma, with the objective of using state-of-the-art nonlinear analysis methods for the design and understanding of nonlinear actuators. The collaboration has already produced a small MATLAB/Octave package freely available at www.ailab.ch/carbajal (section Teaching) under the GPLv3 license.

# 4      Dissemination and special events

*Swiss innovation forum (5. 11. 2009)*

Some of the results and robots developed in the context of this project were presented by M. Lungarella and S. Ravlija at the AI Lab booth at the Swiss Innovation Forum. The Swiss Innovation Forum pools together the knowledge and expertise of the world's leading institutions in the fields of research and innovation.
http://en.ch-innovation.ch

*TV documentary – The robots' intelligence*

The robots and insights developed at the AI Lab and also in the context of this project were presented by Rolf Pfeifer in the documentary "Die Intelligenz der Roboter" (The robots' intelligence), part of the NZZ Format, and broadcasted on the Swiss television (8. 4. 2010).
 http://www.nzzformat.ch/108+M52a6672d686.html

*Vorbild Natur - Wie die Wissenschaft auf neue Idee kommt*

The AI lab was extensively presented in the September issue of Magazin -   Die Zeitschrift des Universität Zürich. Dynamic locomotion and puppy robot developed in the context of this project were also present.
http://www.kommunikation.uzh.ch/publications/magazin/magazin-10-3.html

*Additional lab tours*

Apart from the lab tours that were associated with some of the above-mentioned events, the project was presented to visitors (teachers, grammar school and high school students, representatives from companies, managers, staff from universities of applied science, etc.) in numerous other lab tours (around 20).

# Lectures and invited talks

## Prof. Dr. Rolf Pfeifer

The four messages of embodiment -- how the body shapes the way we think. Hosei University, Tokyo, October 2009.

Embodied intelligence. Invited keynote lecture at the 20th Anniversary of AI in Japan, Nagoya University, October 2009.

Intelligence -- the cooperation of brain, body, and environment. The four messages of embodiment. Chinese Academy of Science, Shanghai, November 2009.

Intelligence -- the cooperation of brain, body and environment. KISTRI -- Kunshan, December 2009.

Können Roboter Denken -- Artificial Intelligence Betweeen Science and Fiction. Private Universität Liechtenstein, Triesen, January, 2010.

Intelligenz -- das Zusammenspiel von Gehirn, Körper und Umwelt. "The four messages of embodiment". Zurich Conference on "Embodiment -- die Intelligenz des Körpers". ETH Zurich, February 2010.

The four messages of embodiment. Robotdoc Lecture, Plymouth, March 2010.

Self-organization, embodiment, and biologically inspired robotics. Invited keynote lecture, International Conference on Chaos Theory. Palermo, Italy, March 2010.

The four messages of embodiment. EU Embodyi Workshop, invited keunote lecture. Livorno, March 2010.

How embodiment changes our view of the mind. Colloquium, Università di Milano-Bicocca. Milano, Italy, April, 2010.

Neurobionik -- das Zusammenspiel von Koerper und neuronaler Informationsverarbeitung. Invited lecture, Neurobionik Wirtschaftsforum. Osnabrueck, Germany, April 2010.

Self-organization, embodiment, and biologically inspired robotics: the four messages of embodiment. Sussex University, UK, April 2010.

Self-organization, embodiment, and biologically inspired robotics: implications for perception and action. Dept. of Machine Perception, Technical University of Prague, June 2010.

Soft robotics. Opening lecture at the International Workshop on Soft Robotics. The University of Tokyo, June, 2010.

Self-organization, embodiment, and biologically inspired robotics. The University of Tokyo, Ikegami Laboratory, June 2010.

Self-organization, embodiment, and biologically inspired robotics. Final lecture (soft robotics workshop) and book launch, The University of Tokyo, June 2010.

Cognition - the interaction of brain, body, and environment. Electrical Engineering Politechnical School, University of Sao Paulo, August 2010.

Generieren und verstehen von Daten. IBM Ittinger Medien-Gespräche, Karthause Ittingen, September 2010.

How embodiment changes our view of the mind (Robotics, embodied intelligence, and philosophy). European Conference on Computing and Philosophy. Munich, September 2010.

Rolf Pfeifer and Matej Hoffmann: Embodiment, morphological computation, and robot body schemas. Body representation workshop, Ascona, Switzerland, September 2010.

Embodiment, morphological computation, and robot body schemas. Monte Verita workshop on body representation. Monte Verita, Ascona, Switzerland, September 2010.

The four messages of embodiment. Invited lecture at the Octopus 2010 Summer School, Livorno, September 2010.

Soft robotics: Self-organization, embodiment, and biologically inspired robotics. Invited plenary lecture. IROS 2010 Conference, Taipei, Taiwan, October 2010.

## Matej Hoffmann

Rolf Pfeifer and Matej Hoffmann: Embodiment, morphological computation, and robot body schemas. Body representation workshop, Ascona, Switzerland, September 2010.

# 5 Overview of the results of the entire project

In this section we briefly summarize the main achievements of the project.[9] First, we will provide an overview of the different solutions to online manipulation of morphology that we have developed. Second, we will briefly recapitulate the results of our investigations on how a robot can autonomously discover, model, and exploit its capabilities. Third, we provide an overview of robotic platforms and simulators that were developed in the context of this project. Finally, we will summarize our activities directed at the educational community.

## (a) Novel actuators and online manipulation of morphology

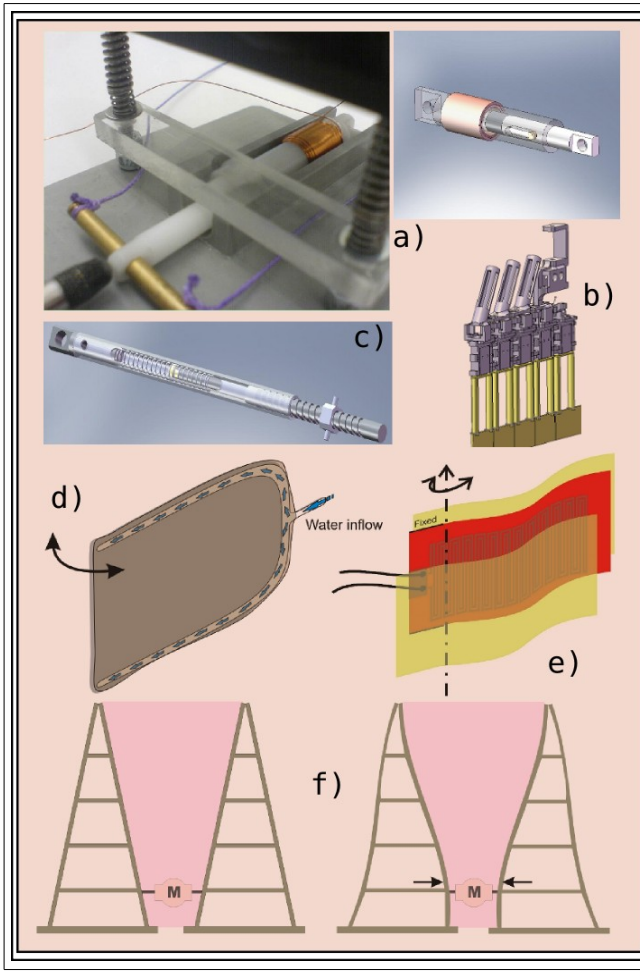

Overview of the actuators developed or tested during the project. In **a)** two views of the magnetic spring, its hardware implementation and its CAD design, which was also used for numerical simulations, is presented. The latest design does not require a coil to produce the same change in force output. Tunable joints for WandaX are shown in **b).** The slots for extension springs are clearly visible; the latest version includes servo motors to control the stiffness of the equivalent rotatory spring. In **c)** design (by E. Benker) of the jackspring device to be used as a tunable tendon for legged robots is shown. We also tested the idea on tunable elastic spines. Several novel actuators for the manipulation of fin stiffness in other swimming platforms were tested achieving moderate success: in **d)** the stiffness of the tail fin was modulated via water pressure on internal tubings; **e)** shows the device tested to vary the elasticity of Polycaprolactone using controlled heat input; **f)** shows the implementation of the Fin Ray Effect® to vary the geometry of the tail fin of a swimming platform. In the scale tested the device did not provide enough force to maintain the modified shape under stress.

## (b) Exploring, modeling, and exploiting the body and its capabilities

The goal is that a robot can automatically explore, learn about, and then exploit the action possibilities it has, given a particular body and environment. These components are very tightly intermingled. Nevertheless, we will structure the overview of our results as follows. First, we will summarize our findings on exploration and exploitation of body dynamics, analyzing various control architectures that can facilitate this goal. Second, we will briefly review our results that build on top of the former – how can a robot model its possible interactions with the environment and use it for anticipation, for instance.

---

[9] We will recapitulate the main results of the project extension only.

**Controllers in tune with body dynamics**

The goal we have set ourselves for a locomotion controller is that it has to discover the capabilities of a particular robot body interacting with its environment and learn to exploit rather than override the 'natural modes' of interaction. We have employed legged and swimming platforms in our investigations.

In a series of Master and Bachelor theses, we were exploring different solutions to this problem in quadrupedal locomotion.[10] In Hutter (2009), and Faessler and Ruegg (2009), we have explored the limits of a feed-forward controller when co-optimized together with the robot morphology. In Nuesch (2009) and Michel (2009), we have investigated two different control architectures, in which oscillators are coupled to the body through feedback connections and, under certain circumstances, get entrained to the resonant frequencies of the body-environment system. In the current project period, we have attempted to integrate these efforts with a modeling effort – a benchmark system for legged locomotion that allows for a comparison of various controller types (Hagmann, 2010).

In the underwater realm, we have concentrated on novel actuator technologies for online manipulation of whole body or tail fin stiffness and on controllers that can exploit this possibility. The details can be taken from this report (WP1 and WP2) and from the publications that accompany it.

**Body schema synthesis and anticipatory behavior**

Whereas in the previous section the body and interaction with the environment were exploited, yet not explicitly modeled, in this section we report our investigations on how a robot can, from its own perspective, develop a model of itself and its action possibilities and how it can employ it to improve its behavior.

First, we were interested in extracting the relations between various sensory and motor modalities. These cross-modal associations are believed to form the basis of cognitive phenomena. Our work in this direction in the quadrupedal as well as swimming (Wanda2.0) platform is still under way. However, we have demonstrated how sensory fusion can be applied in a path integration (dead reckoning) scenario in our quadrupedal platform (see Section "Path integration using multiple sensory modalities" and Reinstein and Hoffmann, 2011).

Second, we have investigated how a quadruped robot can synthesize a forward and inverse model of its gait repertoire. We have developed such a probabilistic model and applied it to a predator-prey scenario. This is an example of future-oriented capabilities in a robot. Please consult the Section "Moving target seeking with prediction and planning" and Oses et al., (2010).

Third, we were working toward integrating our individual case studies and toward abstracting the general principles. To this end we have elaborated a review on body schema in robotics (Hoffmann et al., 2010) and worked on integrating our insights on the implications of embodiment and morphological computation (e.g., Pfeifer and Gomez, 2009, see previous report; Hoffmann and Pfeifer, in preparation).

---

[10]Please refer to the project report from the previous period for details.

## (c) Robots and robot simulators

Overview of the robots built during the execution of the project. **a)** Shows the ZüriHopper, a pneumatic jumping monoped built by E. Benker to study the adaptation to changes in ground stiffness. **b)** Shows the rigid version of the Dumbo robot, Rumbo. The platform was used to understand how direction of motion was affected by the position of the center of mass and the elasticity of the joint, varied using the jackspring. The first modular version of the Puppy robot is shown in **c)**. This platform allowed the use of different legs and bodies in order to study the effects on the motion of the platform. In **d)**, Puppy robot equipped with an inertial measurement unit (white box) is shown. The two versions of the WandaX platform are shown in **e)**. The first version was used to study the role of the passive elasticity of the joint aiming at the creation of the first passive swimmer; the second version included motors to actively change the stiffness online. In **f)** the first autonomous Wanda is shown; its evolved version, Wanda2.0, which was used to study the screw swimming gait, is shown in **g)**.

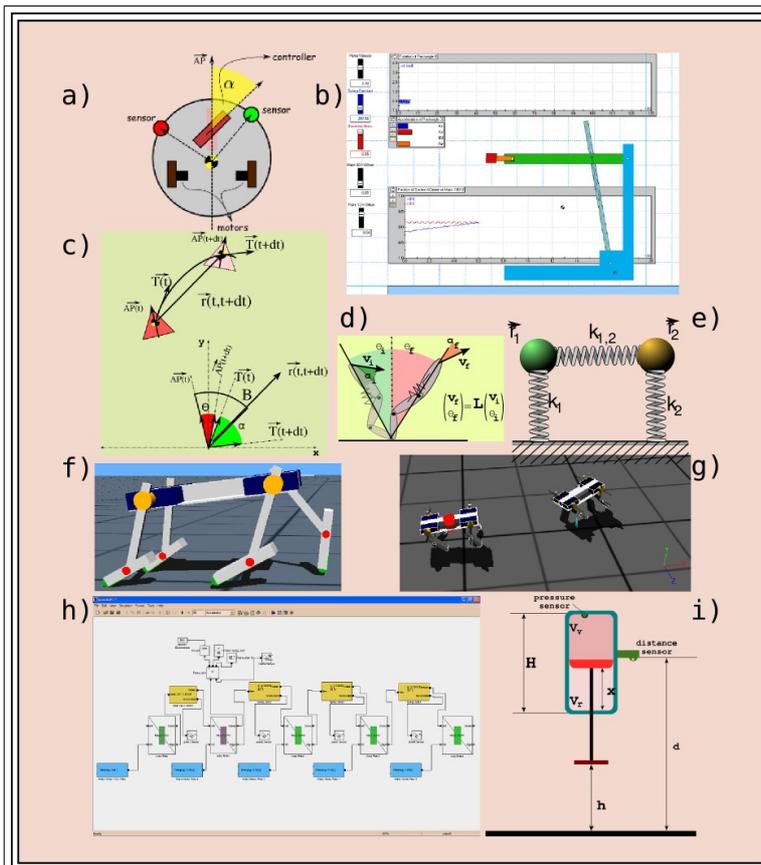

Overview of some of the simulators developed during the project. Light non-holonomic wheeled robots simulators where used to study motion camouflage and modified Braitenberg controllers and are depicted in **a)** and **c)**. These models underly the more sophisticated version used for the Puppies in **g)**. The Rumbo platform in **b)** allowed for the effective creation of simplified models to systematically explore morphological parameters such as center of mass (CoM) position and elasticity of the joint; this simple interactive model provided the insights to understand the effect of CoM position in the direction of motion. The quadruped model in **f)** is a result of morphology optimization for different gaits. The design of the WandaX platform was supported by simulations using SimMechanics®, shown in **h)**, that were later used to model the effect of uncontrolled parameters in the experiment. The pneumatic piston and the jumping dynamics of the Zürihopper **i)** were modeled through ODEs and supported the discovery of the resonance frequencies of the real platform.

## (d) Impact on education

We were also spreading our ideas into the educational community. First, we have taken advantage of the skills and insights learned in the framework of this project in the teaching activities of our laboratory. We want to mention the following courses in this context: (i) Bio-inspired Approaches to Computation and Artificial Intelligence (Fall 2008), Bio-inspired robotics (Spring 2010). In these courses, taught by Dr. Aryananda, the students had the possibility to build their own mobile robots and also take advantage of our quadruped simulation; (ii) Shanghai Lectures (Fall 2009 and 2010). In this global lecture series, Rolf Pfeifer teaches the ideas centered around embodied artificial intelligence to students around the globe; (iii) Marc Ziegler and Matej Hoffmann organized a doctoral seminar on Morphological computation in Spring 2009; (iv) Juan Pablo Carbajal was involved in a course on "Mechatronic systems" taught by Emanuel Benker at DHBW Lorrach in Spring 2010 (Carbajal, Assaf and Benker in preparation) and organized a hands-on workshops on "Dynamical Systems in Matlab" and "Experimental methodologies in robotics: Learning from *older* scientific fields"; (v) The insights partly obtained in this project were also employed in a class for high school computer science teachers (attending the degree program "Master of Advanced Studies Informatics in Upper Secondary Schools") organized by L. Aryananda and D. Assaf in January 2010.

      Second, our framework to bring the fruits of this project to the educational and industrial community is the DREAM project (Development of a Robot kit for Education, Art, and More) running in our laboratory. Within the

16/17

DREAM project, our scientific results are being incorporated into a robotic toolkit that is being developed for educational purposes as well as for researchers as an experimental fast prototyping platform.

# 6 Publication list (last project period, 1. 10. 2009 – 30. 9. 2010)

**Book and Journal**

Hoffmann, M.; Marques, H.; Hernandez Arieta, A.; Sumioka, H.; Lungarella, M. & Pfeifer, R. (2010), 'Body schema in robotics: a review', *IEEE Trans. Auton. Mental Develop.* **(to appear)**.

**Conference Proceedings**

Carbajal, J. P. & Kuppuswamy, N. (2010), Magneto-mechanical actuation model for fin-based locomotion, *in* Design and Nature V: Comparing Design in Nature with Science and Engineering (C. A. Brebbia and A. Carpi, eds.), (Ashurst Lodge, Southampton, UK), pp. 375–387, WIT Press.

Oses, N.; Hoffmann, M. & Koene, R. A. (2010), Embodied Moving-Target Seeking with Prediction and Planning, *in* E. Corchado; M.G. Romay & A.M. Savio, ed.,'Proceeding Hybrid Artificial Intelligence Systems (HAIS), San Sebastian, Spain, Part II', Springer Berlin / Heidelberg.

Reinstein, M. & Hoffmann, M. (2011), Dead reckoning in a dynamic quadruped robot: inertial navigation system aided by a legged odometer, *in* 'Proc. IEEE Int. Conf. Robotics and Automation (ICRA)'. [submitted]

Ziegler, M.; Hoffmann, M. & Pfeifer, R. (2011), Varying body stiffness for aquatic locomotion, *in* 'Proc. IEEE Int. Conf. Robotics and Automation (ICRA)'. [submitted]

**Publications in preparation**

Carbajal, J. P.; Assaf, D. & Benker, E. Using robots to engage students in scientific thinking. [in preparation: to be submitted to the American Association of Physics Teachers.]

Hoffmann, M. & Pfeifer, R. The implications of embodiment for behavior and cognition: animal and robotic case studies. [in preparation]

Ziegler, M. & Pfeifer, R. Expanding the behavior repertoire through manipulation of tail fin stiffness in a swimming robot. [in preparation]

**Master thesis**

Hagmann, E. (2010), A simplified approach towards legged locomotion control. Unpublished Master thesis, ETH Zurich, Switzerland.